\documentclass[11pt]{article}
\usepackage{coling2018}
\usepackage{times}
\usepackage{url}
\usepackage{latexsym}
\usepackage{color}

\usepackage{color}

\usepackage{tabularx}
\usepackage{footnote}
\usepackage{graphicx}
 \usepackage{multirow}

\usepackage{CJKutf8}
\usepackage[russian,english]{babel}

\slname{Russian}

\title{All-in-one: Multi-task Learning for Rumour Verification  }

\sltitle{

\begin{CJK}{UTF8}{gbsn} \foreignlanguage{russian}{ Три в одном: Многозадачное Обучение для Верификации Слухов} \end{CJK}

}

\author{
Elena Kochkina$^{1,2}$, Maria Liakata$^{1,2}$, Arkaitz Zubiaga$^1$\\
       $^1$ University of Warwick, Coventry, United Kingdom\\
       $^2$ Alan Turing Institute, London, United Kingdom\\
       \{E.Kochkina, M.Liakata, A.Zubiaga\}@warwick.ac.uk
}

\date{}
\begin{document}
\maketitle
\begin{abstract}
Automatic resolution of rumours is a challenging task that can be broken down into smaller components that make up a pipeline, including rumour detection, rumour tracking and stance classification, leading to the final outcome of determining the veracity of a rumour. In previous work, these steps in the process of rumour verification have been developed as separate components where the output of one feeds into the next. We propose a multi-task learning approach that allows joint training of the main and auxiliary tasks, improving the  performance of rumour verification. We examine the connection between the dataset properties and the outcomes of the multi-task learning models used.
\end{abstract}

\makesltitle
\begin{slabstract}
\begin{CJK}{UTF8}{gbsn}
 
 \foreignlanguage{russian}{Автоматическая верификация слухов в Интернете - это сложная задача, которая может быть разбита на подзадачи, включающие в себя обнаружение слухов, отслеживание слухов, классификацию отношения пользователей к правдивости этих слухов и, в итоге, разрешение основной задачи - определение достоверности слуха.
 В предыдущих исследованиях модели для решения каждой из подзадач разрабатывались как отдельные компоненты, где вывод модели для задачи предыдущего этапа являлся вводдом для модели задачи следующего этапа.
 Мы предлагаем использование многозадачного обучения для совместного обучения основной и вспомогательных задач в одной модели, что позволяет улучшить качество модели верификации слухов. Также мы рассматриваем связь между свойствами набора данных и результатами использования моделей с многозадачным обучением.} 
 
\end{CJK}
\end{slabstract}

\section{Introduction}
\label{intro}
\blfootnote{
    %
    %
     \hspace{-0.65cm}  
     This work is licenced under a Creative Commons 
     Attribution 4.0 International Licence.
     Licence details:
     \url{http://creativecommons.org/licenses/by/4.0/}
     
    %
}

Social media have gained popularity as platforms 
that enable users to follow events and breaking news \cite{hu2012breaking}. 
However, not all information that spreads on social media during such events is accurate.
The serious harm that inaccurate information can cause to society in  critical situations \cite{lewandowsky2012misinformation} has led to an increased interest within the scientific community to develop tools to verify information from social media \cite{shu2017fake}. Likewise, social media platforms themselves, such as Facebook,\footnote{https://www.recode.net/2017/2/16/14632726/mark-zuckerberg-facebook-manifesto-fake-news-terrorism} are investing significant effort in mitigating the problems caused by misinformation. 
One of the features that characterises social media 
is the rapid emergence and spread of new information.
This leads to the circulation of rumours, claims that are unverified at the time of posting, and for which eventually evidence proving their true or false nature may come out \cite{difonzo2007rumor}. Therefore, a rumour resolution system needs to go through a set of steps, from detecting that a new circulating claim is a rumour, to the ultimate step of determining its veracity value.

The rumour resolution process has been defined as a pipeline involving four sub-tasks \cite{zubiaga2017detection} (see  Figure \ref{fig:pipe}): (1) rumour detection, determining whether a claim is worth verifying rather than the expression of an opinion; (2) rumour tracking, collecting sources and opinions on a rumour as it unfolds; (3) stance classification, determining the attitude of the sources or users towards the truthfulness of the rumour, and (4) rumour verification, as the ultimate step where the veracity value of the rumour is predicted. These steps can be performed at different times in the life-cycle of a rumour, making this a time-sensitive process. Ideally, rumours can be resolved as either true or false. However, they can also remain unverified when there is no sufficient evidence to determine their veracity \cite{caplow1947rumors}.

A recent body of work studies each of these four tasks separately \cite{lukasik2016hawkes,liu2016reuters,enayet2017niletmrg}. However, the way these subtasks interact and their integration into a complete rumour resolution system is yet to be explored. In this work we assume that veracity classification is the crucial component of a rumour resolution system, as the final output that determines, given information collected about a rumour at a certain point in time, if the rumour is true, false, or remains unverified. We express the rumour resolution process as a multi-task problem that needs to address a number of challenges, where the veracity classification task is the main task and the rest of the components are auxiliary tasks that can be leveraged to boost the performance of the veracity classifier.

We propose to achieve this setting by using a multi-task learning approach. Multi-task learning \cite{caruana1998multitask} refers to the joint training of multiple tasks, which has gained popularity recently for a range of tasks in Machine Learning and Natural Language Processing \cite{collobert2008unified} and has been applied in a number of different tasks and machine learning architectures. Its effectiveness is mainly attributed to learning shared representations of closely related tasks, such that two complementary tasks can give each other `hints'.

We propose joint learning of the tasks in the verification pipeline, that will allow us to leverage relations between the tasks.  We assess the effectiveness of using a multi-task learning approach for the rumour resolution process in four different scenarios: (1) performing single task learning to perform veracity classification, (2) performing multi-task learning that combines stance and veracity classification with the aim of boosting performance of the latter,
(3) performing multi-task learning that combines rumour detection and verification, comparing the improvement brought by each of the two auxiliary tasks, and (4) performing multi-task learning that combines rumour detection, stance classification and veracity prediction to improve the performance for veracity. We employ a deep learning architecture that views each subtask as having a sequential nature. We compare our results with a state-of-the-art veracity classification approach introduced by Enayet  and  El-Beltagy \shortcite{enayet2017niletmrg}.

While a lot of work reports positive outcomes of the application of multi-task learning to various NLP tasks \cite{collobert2008unified,aguilar2017multi,lan2017multi}, there are also studies showing that this is not always the case \cite{alonso2017multitask,bingel2017identifying}. Alonso and Plank \shortcite{alonso2017multitask} were the first to demonstrate that multi-task learning brings benefits only for some combinations of main and auxiliary tasks. They also investigate the relationship between the multi-task learning outcome and the properties of the dataset. We perform a similar analysis to examine the link between the properties of our rumour datasets and the results of the multi-task learning approach used. 

Our results show that a multi-task learning scenario that leverages all three subtasks, where veracity classification is the main task and  stance classification and rumour detection are auxiliary tasks, leads to substantial improvements over a standard, single task veracity classification system, as well as a majority baseline and a state-of-the-art system. The combination of all three subtasks also outperforms the multi-task learning scenario where only two of the subtasks are combined.
 
\section{Related work}
\label{sec:relevant}
\subsection{Rumour classification system}
\begin{figure}[h]
 \vspace{-1cm}
    \centering
    \includegraphics[width=0.9\textwidth]{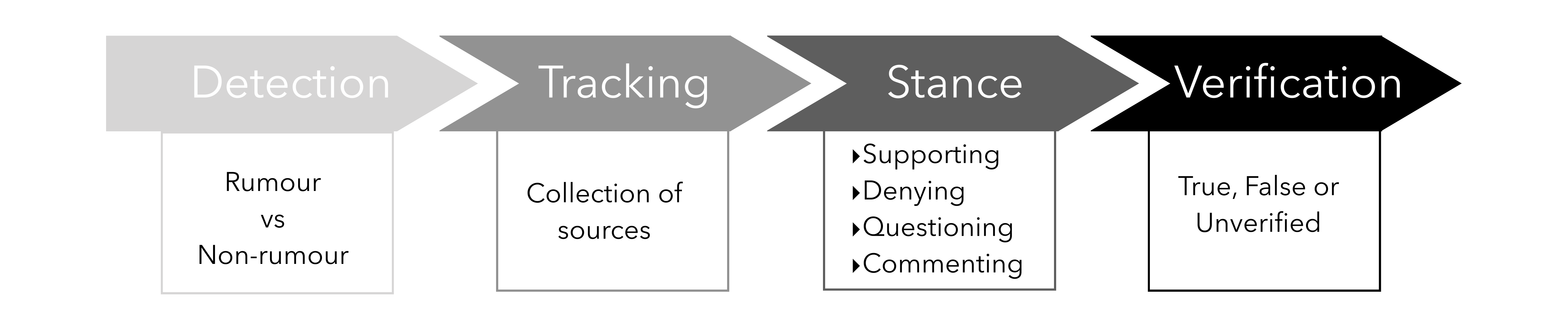}
    \vspace{-0.5cm}
    \caption{Rumour resolution pipeline.}
    \label{fig:pipe}
\end{figure}
Rumour classification is a complex task that can be represented as a pipeline of sub-tasks. While there are different possibilities for structuring this pipeline, here we adopt the architecture of a rumour classification system proposed in Zubiaga et al. \shortcite{zubiaga2017detection} (see Figure \ref{fig:pipe}). Thus, a rumour classification system is expressed as a sequence of subtasks, namely rumour detection, rumour tracking, rumour stance classification leading to rumour verification. In this work we consider three of these tasks: rumour detection, stance and verification, focusing on the evaluation of the latter. Rumour tracking does not involve classification but rather consists of collecting tweets following up on a rumour in the form of replies.

A recent survey \cite{zubiaga2017detection} defines rumour detection  as the task which distinguishes rumours (unverified pieces of information) from non-rumours (all other circulating information). Here we adhere to this definition. Therefore, once rumour detection is performed, the information classified as a rumour is then fed into the the stance and veracity classification components of a system to ultimately determine the veracity of the rumour.

While we are not aware of previous work tackling the entire rumour classification pipeline, there has been a body of work tackling each of the tasks individually.

\paragraph{Stance classification.}
An increasing body of work has focused on the stance classification component, i.e. determining whether different posts associated with a rumour support it, deny it, query it, or just comment on it. 
The stance expressed by users towards a particular rumour can be indicative of the veracity of a rumour, furthermore, 
it has been shown that rumours attracting higher levels of skepticism in the form of denials and questioning responses are more likely to be proven false later \cite{mendoza2010twitter,procter2013readinga,derczynski2014pheme}. Previous work on stance classification \cite{lukasik2016hawkes,kochkina2017turing,zubiaga2017exploiting,zubiaga2018discourse} has explored the use of sequential classifiers, treating the task as one that evolves over time; it has been shown that these sequential classifiers substantially outperform standard, non-sequential classifiers. 

\paragraph{Rumour detection.} Work on rumour detection is more scarce. One of the first approaches was introduced by Zhao et al. \shortcite{zhao2015enquiring}, who built a rule-based approach to identify skepticism (e.g. is this true?) and therefore determine that the associated information is a rumour. The limitations of this approach consist in having to wait for responses to arrive, as well as lack of generalisability due to manually-defined rules. Zubiaga et al \shortcite{zubiaga2017exploiting} proposed a sequential approach to leverage context from earlier posts during an event. The sequential approach achieved significant improvements, especially in terms of recall, where the rule-based approach proved limited.

\paragraph{Rumour verification.} Common approaches to rumour verification involve first the collection of corpora of resolved rumours from rumour debunking websites such as snopes.com, emergent.com, politifact.com. Wang \shortcite{wang2017liar} created a dataset based on claims from politifact.com that are annotated for degrees of truthfulness. To resolve these they propose a hybrid convolutional
neural network that integrates metadata with text. Twitter is a popular platform to study rumours, while often seeds and annotations for rumours are still taken from rumour debunking websites. Giasemidis et al. \shortcite{giasemidis2016determining} collected a dataset of 72 rumours from Twitter. This work measured trustworthiness of a claim at varying time windows. Boididou et al. \shortcite{boididou2014challenges,boididou2017learning} have focused on tweets that are related to fake images, bringing in the multimedia component into the verification process. 

\paragraph{Sequence classification.} Works that consider tasks of rumour detection and verification (within their own definition of the tasks, not necessarily aligning with the one used here) also highlighted the importance of a sequential time-sensitive approach when dealing with rumours  \cite{ma2016detecting,kwon2017rumor,chen2017call}. As previous work in the literature has highlighted the importance of sequential classifiers in rumour detection we have decided to use an LSTM based architecture for our experiments.

\begin{figure}[h]
\vspace{-0cm}
    \centering
    \includegraphics[width=0.8\textwidth]{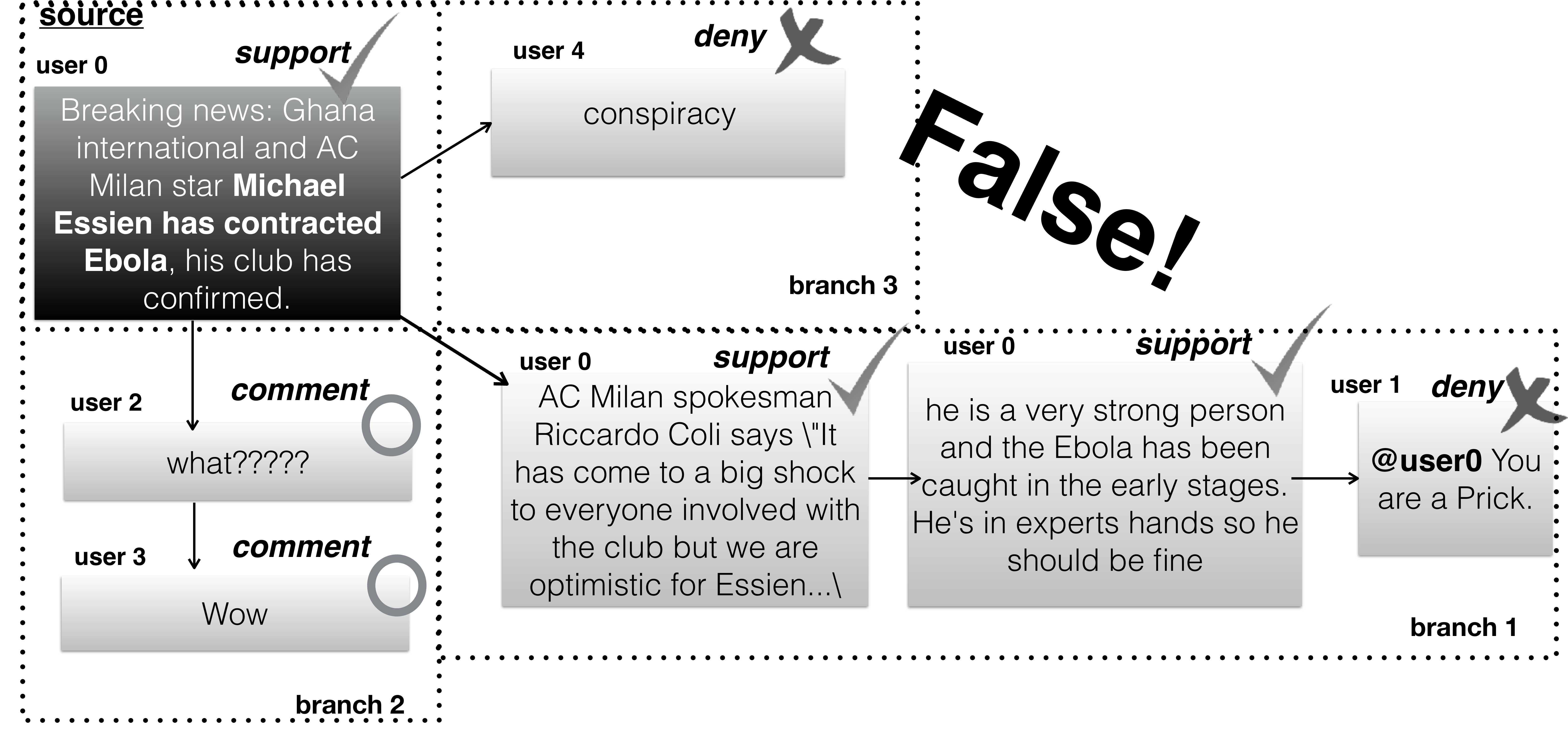}
    \caption{Example of a conversation with 3 branches from the RumourEval dataset.}
    \label{fig:conv}
\end{figure}

\subsection{Multi-task learning}
Multi-task learning refers to the joint learning of several related tasks with a shared representation. In recent years, a multi-task learning approach has been successfully applied in combination with neural networks for a variety of NLP tasks \cite{collobert2008unified}.
In this work we use the most common approach to multi-task learning, namely hard parameter sharing, meaning that different tasks are using the same hidden layer(s).
The effectiveness of a multi-task learning approach is attributed to: effectively increasing the size of the training set by using additional datasets for related tasks and regularisation, as the model has to learn a shared representation for multiple tasks there is less risk of overfitting on one of them.

In multi-task learning auxiliary tasks can be used to direct the main task to use/learn the features that it otherwise would have ignored or failed to identify due to complex relations between the tasks and features. For example, multi-task learning is particularly useful when potentially helpful features are not used as such for the main task, but become labels in an auxiliary task. This use case is very relevant to this work; stance classification could be used as a feature in a system for veracity classification, as has indeed been the case in previous studies, which have shown a relationship between the two tasks. Zhao et al. \shortcite{zhao2015enquiring} and Enayet  and  El-Beltagy \shortcite{enayet2017niletmrg} have created successful models using this premise. However these studies assume access to stance and veracity labels for the same data,  which does not apply in our case as we do not have stance labels for all of the threads in the dataset (see section \ref{sec:data}). 

\section{Data}
\label{sec:data}

We use two publicly available datasets of rumours: PHEME \cite{zubiaga2016analysing,zubiaga2017exploiting} and RumourEval \cite{derczynski2017semeval} that contain different levels of annotation for the tasks of rumour detection, rumour stance and veracity classification. 
Both datasets  contain  Twitter  conversation  threads  associated with different newsworthy events including the Ferguson unrest, the shooting at Charlie Hebdo, the shooting in Ottawa, the hostage situation in Sydney and the crash of a Germanwings plane.
Figure \ref{fig:conv} shows an example of a conversation discussing a rumour about Michael Essien having contracted Ebola. The conversation consists of a source tweet conveying a rumour and a tree of responses, expressing their opinion towards the claim contained in the source tweet. The veracity label of the rumour is false, while each of the responses could be tagged as either supporting, denying, questioning or commenting on the rumour. Conversations can be decomposed into branches, such that a branch is a linear sequence of tweets starting from a leaf node of the conversation tree and going through its parent nodes up to the source tweet. The conversation on the Figure \ref{fig:conv} can be decomposed into three branches.

\subsection{RumourEval}
RumourEval is a dataset that was released as part of the SemEval-2017 Task 8 competition \cite{derczynski2017semeval}. It contains 325 Twitter threads discussing rumours. All conversations in the RumourEval dataset are rumours, therefore this dataset only covers the tasks of rumour stance and veracity classification. It is split into training, testing and development sets. The testing set contains a mix of rumours related to the same events as in the training and development sets, with the addition of two rumours: about Marina Joyce and the health condition of Hillary Clinton. Table \ref{tab1} shows the number of conversation threads, branches and tweets in each of the sets of the RumourEval dataset, as well as the class distribution for both tasks. In the stance classification task there is a strong class imbalance towards the commenting class, while questioning and denying are minority classes, which holds true for all subsets. For the verification task, the training set contains more true instances than false or unverified, whereas the development and testing sets are more balanced. 
\begin{table}[t]
\centering
\resizebox{0.9\textwidth}{!}{%
\begin{tabular}{|l|l|l|l|l|l|l|l|l|l|l|}
\hline
               & \textbf{Threads} & \textbf{Branches} & \textbf{Tweets} & \textbf{True} & \textbf{False} & \textbf{Unverified} & \textbf{S} & \textbf{D} & \textbf{Q} & \textbf{C} \\ \hline
Development    & 25                 & 215                  & 281                & 10            & 12             & 3                   & 69         & 11         & 28         & 173        \\ \hline
Testing        & 28                 & 772                  & 1049               & 8             & 12             & 8                   & 94         & 71         & 106        & 778        \\ \hline
Training       & 272                & 3030                 & 4238               & 127           & 50             & 95                  & 841        & 333        & 330        & 2734       \\ \hline
\textbf{Total} & 325                & 4017                 & 5568               & 145           & 74             & 106                 & 1004       & 415        & 464        & 3685       \\ \hline
\end{tabular}
}
\caption{Number of threads, tweets and class distribution in RumourEval dataset.}
\label{tab1}
\end{table}
\subsection{PHEME}
We are making available the extended version of the PHEME dataset of rumours and non-rumours related to nine events, each updated with veracity information for each of the rumours\footnote{\url{https://figshare.com/articles/PHEME_dataset_for_Rumour_Detection_and_Veracity_Classification/6392078}}.

This dataset contains 3 levels of annotation. First, each thread is annotated as either rumour or non-rumour; second,  rumours are labeled as either true, false or unverified. And third, a subset (threads used in RumourEval) is annotated for stance classification at the tweet level through crowd-sourcing.

Rumours in this dataset were labeled as true, false and unverified by professional journalists \cite{zubiaga2016analysing}. While stance can be labeled by non-expert workers as labels can be inferred directly from the text, the rumour verification task is more challenging as it requires analysis of the context, and further understanding 
of the rumours
 in order to determine if the underlying story is true, false, or remains unverified. 
To assess the difficulty of performing the verification task by a non-expert, an author of this paper went through the rumours and annotated them for veracity. The overlap between the non-expert annotator and the journalist was within the range of $60-65\%$ on the rumour stories from five largest events.

Table \ref{tab2} shows the size of each event in the PHEME dataset as well as the label distribution for the tasks of rumour detection and verification. The information about the stance classification task is in table \ref{tab1}. Events differ in size drastically and have different class label proportions. Overall, the PHEME dataset contains fewer rumours than non-rumours, while the majority class for rumours is true. 
\begin{table}[t]
\centering
\resizebox{0.9\textwidth}{!}{%
\begin{tabular}{|l|l|l|l|l|l|l|l|}
\hline
\textbf{Events}   & \textbf{Threads} & \textbf{Tweets} & \textbf{Rumours} & \textbf{Non-rumours} & \textbf{True} & \textbf{False} & \textbf{Unverified} \\ \hline
Charlie Hebdo     & 2,079               & 38,268             & 458                 & 1,621                   & 193              & 116               & 149                    \\ \hline
Sydney siege      & 1,221               & 23,996             & 522                 & 699                     & 382              & 86                & 54                     \\ \hline
Ferguson          & 1,143               & 24,175             & 284                 & 859                     & 10               & 8                 & 266                    \\ \hline
Ottawa shooting   & 890                 & 12,284             & 470                 & 420                     & 329              & 72                & 69                     \\ \hline
Germanwings-crash & 469                 & 4,489              & 238                 & 231                     & 94               & 111               & 33                     \\ \hline
Putin missing     & 238                 & 835                & 126                 & 112                     & 0                & 9                 & 117                    \\ \hline
Prince Toronto    & 233                 & 902                & 229                 & 4                       & 0                & 222               & 7                      \\ \hline
Gurlitt           & 138                 & 179                & 61                  & 77                      & 59               & 0                 & 2                      \\ \hline
Ebola Essien      & 14                  & 226                & 14                  & 0                       & 0                & 14                & 0                      \\ \hline
\textbf{Total}    & 6,425               & 105,354            & 2,402               & 4,023                   & 1,067            & 638               & 697                    \\ \hline
\end{tabular}
}
\caption{Number of threads, tweets and class distribution in the PHEME dataset.}
\label{tab2}
\end{table}

We perform two types of experiments using different subsets of the PHEME dataset: (1) using the five largest events (see Table \ref{tab2}) and (2) using all nine events. The five largest events create a more balanced dataset as those are major crisis events during which true updates and false information on different aspects of the event were shared and discussed, whereas the 4 smallest events only contain a single rumour story at the core of the event. In both cases, cross-validation experiments are performed by relying on a leave-one-event-out setting, i.e. using all the events except the one left for testing in each case. This set up is more challenging than the one presented in RumourEval, however it is more representative of the real world. We micro-average the performance across events.

\section{Models}

\subsection{Sequential approach}
As the benefits of using a sequential approach were suggested by previous studies of rumour stance classification and rumour detection tasks (discussed in Section \ref{sec:relevant}), we are following the branchLSTM approach described in Zubiaga et al.  \cite{zubiaga2018discourse}. We split the conversations into linear branches and use them as training instances that become an input to a model consisting of an LSTM layer followed by several dense ReLU layers and a softmax layer that predicts class probabilities. 
As the stance classification task is annotated at the tweet level, we use the output from each time step of LSTM, whereas for the tasks of rumour detection and verification, annotated at the thread level, we are only using outputs from the final time steps (this idea is illustrated in Figure \ref{fig:models}). To get per-thread predictions, we use majority voting for each of the branches from the thread. 
The model is trained using categorical cross entropy loss.

\subsection{Multi-task learning approach}
\begin{figure}[t]
    \centering
    \includegraphics[width=0.7\textwidth]{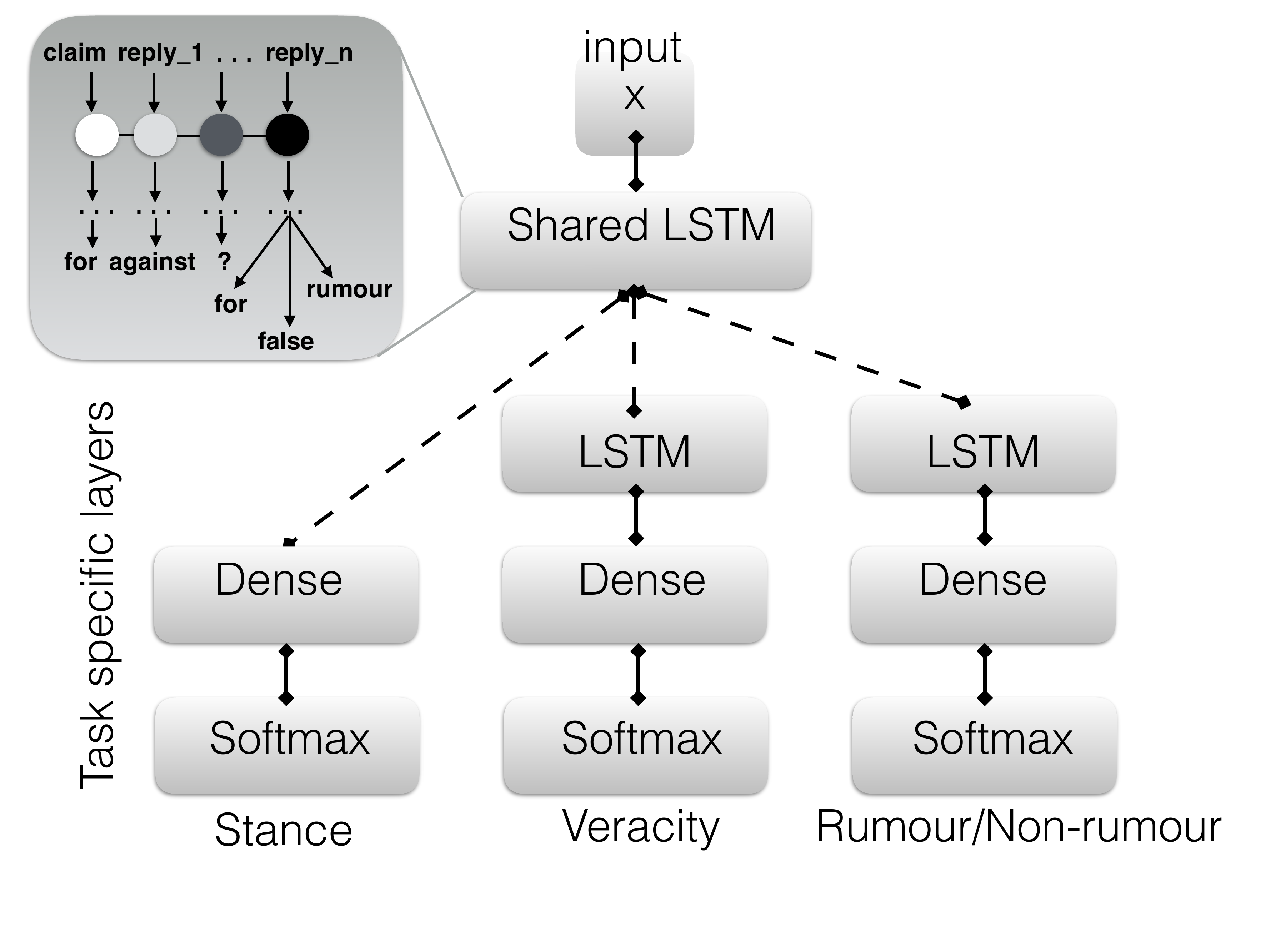}
    \vspace{-0.5cm}
    \caption{Multi-task learning models. Dotted lines represent that same set up is used for learning one, two or three tasks. Input format is a branch of tweets. Stance classification implies predictions per-tweet, whereas detection and verification tasks only require per-branch output.}
    \label{fig:models}
    \vspace{-0.3cm}
\end{figure}

We leverage the relationship between the tasks from the rumour classification pipeline in a joint multi-task learning setup. Figure \ref{fig:models} illustrates our approach.  At the base of it is a sequential approach, as discussed above, represented by a shared LSTM layer (hard parameter sharing), which is followed by a number of task-specific layers. 
The possible task combinations are shown as dotted lines on Figure \ref{fig:models} that can be present or absent depending on the combination.  We perform experiments in three set ups: joint training of (1) stance or (2) rumour detection together with veracity classification, and (3) learning all three tasks together.
The cost function in the multi-task models  is a sum of losses from each of the tasks. Datasets for each of the three tasks are not equal in size, therefore when the training instance is lacking a label for one of the tasks, its prediction does not add anything to the loss function, as if it had been predicted correctly. 

\subsection{Baselines}
We compare proposed sequential and multi-task learning approaches with several baselines. First of all, majority vote, a strong baseline which results in high accuracy due to the class imbalance in the veracity classification task.
Another strong benchmark is NileTMRG \cite{enayet2017niletmrg}, the best veracity classification system from SemEval-2017 Task 8. The NileTMRG model is based on a linear SVM that uses a bag-of-words representation of the tweet concatenated with selected features: presence of URL, presence of hashtag and proportion of supporting, denying and querying tweets in the thread. 
We have made our own implementation NileTMRG* of the NileTMRG model based on their SemEval paper. 
This model requires stance labels for each of the tweets in the dataset, however these are not available for the PHEME dataset. To obtain the proportion of different stance labels among responses in the thread, we use our own implementation of the stance classification model from Kochkina et al. \shortcite{kochkina2017turing}. On the RumourEval dataset our implementation NileTMRG* achieves better results than the original  NileTMRG score \cite{enayet2017niletmrg} (accuracy 0.570 vs reported 0.536) as it utilises a better model for stance classification. 
The NileTMRG model is a baseline showing the scenario where pipeline tasks are performed sequentially and the outcome of the previous step (stance classification) is an input to the next one (veracity classification). The NileTMRG model has set a precedent in demonstrating that patterns of support are useful indicators of rumour veracity. 

\subsection{Features}
We perform the following pre-processing of the tweets in the datasets: remove nonalphabetic characters, convert all words to lower case and tokenise texts\footnote{For implementation of all pre-processing routines we use Python 2.7 with the NLTK package}. Once tweet texts are pre-processed, we extract word2vec word embeddings pre-trained on the Google
News dataset (300d)\footnote{Embeddings were retrieved from \url{https://code.google.com/archive/p/word2vec/}. Processing was performed using gensim package \cite{rehurek_lrec}. } \cite{mikolov2013efficient} for each word in a tweet and take the average, thus obtaining a tweet representation. This representation was shown to work well for tweets due to their short length \cite{kochkina2017turing}.

\section{Experiment setup}

\subsection{Hyperparameters}

 We determined the optimal set of hyperparameters 
 by testing the performance of our models on the development set for different parameter combinations. We used the Tree of Parzen Estimators (TPE) algorithm\footnote{We used the implementation of the TPE algorithm in the hyperopt package} to search the parameter space and minimise the loss function expressed as $(1-macro F)$ for single task, and $(1-macroF_a)(1-macroF_b)$ or $(1-macroF_a)(1-macroF_b)(1-macroF_c)$ for multi-task learning of two and three tasks respectively. This loss function gives equal weight to all tasks.
The parameter space is defined as follows: the number of dense ReLU layers varies from one to four; the number of LSTM layers is \{1, 2\}; the mini-batch size is 32; the number of units in the ReLU layer is \{300, 400, 500, 600\}, and in the LSTM layer is \{100, 200, 300\}; the strength of the L2 regularisation is \{$10^{-4}$, $10^{-3}$\} and the number of epochs is 50.  We performed 30 trials of different parameter combinations optimising for accuracy on the development set in order to choose the best combination. We also use 50\% dropout before the output layer.
Zero-padding and masks that account for the varying lengths of the input branches were used in all our models. Models were implemented\footnote{https://github.com/kochkinaelena/Multitask4Veracity} using Python 3 and the Keras package. 

\subsection{Evaluation}

The RumourEval dataset was provided with a training/development/testing split. We tune parameters on the development set and then retrain the model on the combined training and development sets before evaluation on the testing set. 
On the PHEME dataset we perform leave-one-event-out (LOEO) cross-validation, which makes this task set up harder than for RumourEval but closer to the realistic scenario where we want to verify unseen rumours. When choosing parameters, the Charlie Hebdo event was used as the development set as it has balanced labels. 
We evaluate models using accuracy and macro-averaged F-score as the tasks in the PHEME dataset suffer from a class imbalance. 

\section{Results and Discussion}

The main results of our experiments are presented in Table \ref{tab:compare4}. It shows the results of the multi-task learning models MTL2 with two tasks: Veracity+Stance and Veracity+Detection; MTL3 with three tasks Stance+Veracity+Detection, Majority, NileTMRG* and single task branchLSTM baselines on the main task, veracity classification. 

As the datasets contain a significant class imbalance, the majority baseline achieves fairly high accuracy scores. However due to the nature of this task, it is more important for a model to recognize all of the classes, especially false rumours, therefore the macro-averaged F-score is more important for performance evaluation. All models demonstrate improvement over the majority baseline in terms of macro F-score. We observe improvements of multi-task approaches over single task learning in both accuracy and macro F-score, and adding the third task brings further improvement. 

MTL2 Veracity+Detection and MTL3 experiments were performed only on the PHEME dataset as the RumourEval dataset consists only of rumours (no rumour detection). On the RumourEval dataset the single task model branchLSTM outperforms the majority baseline, although it does not perform as well as NileTMRG*, while MTL2 shows improvement over both NileTMRG* and branchLSTM.
Experiments on the PHEME dataset also show a pattern of increasing scores: MTL2 outperforms single task models and MTL3 outperforms MTL2. 
\begin{table}[t]
\centering
\resizebox{\textwidth}{!}{%
\begin{tabular}{|l|l|l|l|l|l|l|}
\hline
\multicolumn{7}{|c|}{\textbf{RumourEval}}                                                                          \\ \hline
         & \textbf{Majority (True)} & \textbf{NileTMRG*}  & \textbf{branchLSTM} & \textbf{\shortstack{MTL2 \\ Veracity+Stance}} & \textbf{\shortstack{MTL2 \\ Veracity+Detection}} & \textbf{MTL3} \\ \hline
Macro F  & 0.148                    & 0.539                  & 0.491                & 0.558  & -       & -             \\ \hline
Accuracy & 0.286                    & 0.570              & 0.500                  & 0.571  & -        & -             \\ \hline
\multicolumn{7}{|c|}{\textbf{PHEME 5 events}}                                                                   \\ \hline
         & \textbf{Majority (True)} & \textbf{NileTMRG*} & \textbf{branchLSTM} & \textbf{\shortstack{MTL2 \\ Veracity+Stance}} &\textbf{\shortstack{MTL2 \\ Veracity+Detection}} & \textbf{MTL3} \\ \hline
Macro F  & 0.226                    & 0.339              & 0.336                & 0.376     & 0.373    & 0.396         \\ \hline
Accuracy & 0.511                     & 0.438              & 0.454                & 0.441    & 0.410      & 0.492         \\ \hline
\multicolumn{7}{|c|}{\textbf{PHEME 9 events}}                                                                   \\ \hline
         & \textbf{Majority (True)} & \textbf{NileTMRG*} & \textbf{branchLSTM} & \textbf{\shortstack{MTL2 \\ Veracity+Stance}} & \textbf{\shortstack{MTL2 \\ Veracity+Detection}} & \textbf{MTL3} \\ \hline
Macro F  & 0.205                    & 0.297              & 0.259                & 0.318   & 0.345      & 0.405         \\ \hline
Accuracy & 0.444                    & 0.360              & 0.314                & 0.357   & 0.397      & 0.405         \\ \hline
\end{tabular}
}
\caption{Comparison of performance of sequential single task approach, multi-task learning approaches with two and three tasks with majority and NileTMRG baselines on veracity classification task.}
\label{tab:compare4}
\end{table}
Comparing the performance of the MTL2 model using stance as an auxiliary task with the model using rumour detection as an auxiliary task on the PHEME 5 events dataset we observe that both models bring an improvement over the single task branchLSTM baseline. 

When using 9 events of the PHEME dataset we observe worse performance than on 5 events. Even though we are adding more training data, the 4 additional events are qualitatively different to the 5 large news-breaking events. Each of these 5 large events contained rumours labeled with all classes as well as non-rumours, whereas the 4 additional events are small and the event itself is a false or unverified rumour. This highlights the difficulty of the rumour verification task in a leave-one-event-out setup and the importance of high quality data.
NileTMRG* is a very strong baseline, and while the single task branchLSTM model is competitive when we are using 5 largest events, NileTMRG* is only outperformed by the multi-task learning approach. 

\subsection{Per-event and per-class results analysis}
\begin{figure}[t]
    \centering
    \includegraphics[width=\textwidth]{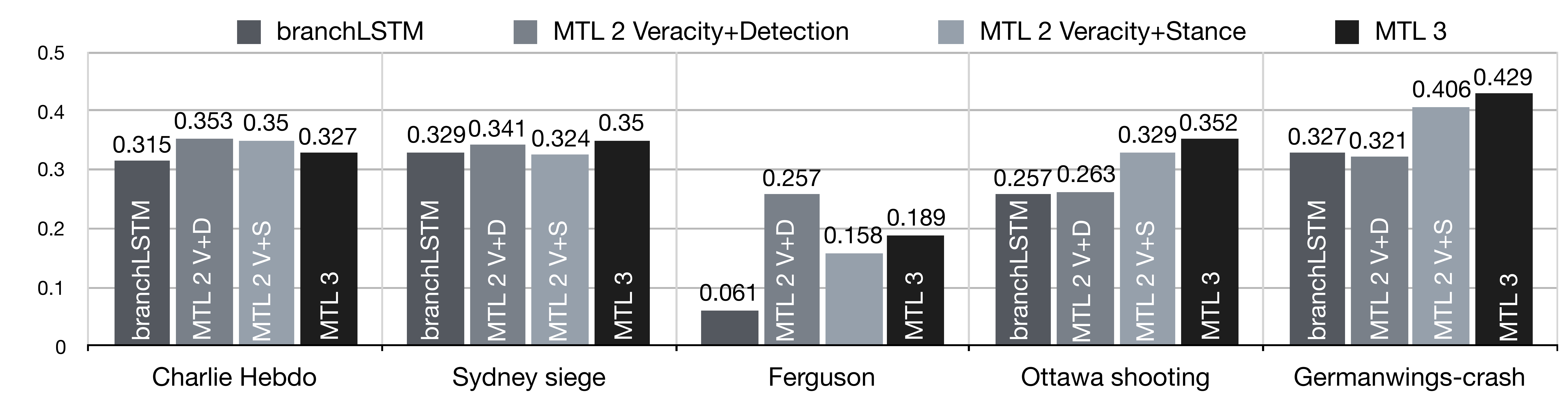}
    \caption{Comparison of macro F-score of sequential single task approach (branchLSTM), multi-task learning approaches with two and three tasks on different events from the PHEME dataset.}
    \label{fig5}
\end{figure}

Here we analyse the performance of proposed models on each of the 5 largest events. Figure \ref{fig5} illustrates the comparison of macro-averaged F-scores of the proposed models for each of the events. Multi-task learning models outperform single task learning approaches for each event. The Ferguson event is the hardest one for all of the models as it has a different class distribution to all other events (see Table \ref{tab2}). 

Table \ref{tab:event} shows per event and per class performance of the multi-task learning model that incorporates all three tasks (MTL3). We have analysed similar performance breakdown tables for other models but omit them here because of space constraints. All models tend to predict the majority class (true) the best. As the Ferguson event is strongly dominated by unverified rumours, it is the only event with high performance on the unverified class. Single task models (NileTMRG*, branchLSTM) are better at identifying false than unverified rumours, whereas multitask models (MTL2, MTL3) are better at  identifying unverified than false rumours. 
\begin{table}[t]
\centering
\resizebox{0.7\textwidth}{!}{%
\begin{tabular}{|l|l|l|l|l|l|}
\hline
MTL3 5 events               & \textbf{MacroF} & \textbf{Accuracy} & \textbf{TRUE} & \textbf{FALSE} & \textbf{Unverified} \\ \hline
\textbf{Charlie Hebdo}     & 0.327           & 0.369             & 0.502         & 0.227          & 0.251               \\ \hline
\textbf{Sydney siege}      & 0.350            & 0.575             & 0.731         & 0.153          & 0.168               \\ \hline
\textbf{Ferguson}          & 0.189           & 0.338             & 0.058         & 0              & 0.508               \\ \hline
\textbf{Ottawa shooting}   & 0.352           & 0.645             & 0.789         & 0.168          & 0.100                 \\ \hline
\textbf{Germanwings-crash} & 0.429           & 0.420              & 0.538         & 0.358          & 0.364               \\ \hline
\end{tabular}
}
\caption{Per event and per-class results for multi-task learning approach with 3 tasks on PHEME 5 events.}
\label{tab:event}
\end{table}

\subsection{Analysis of data properties}
Alonso and Plank \shortcite{alonso2017multitask} link the gains/losses in performance from multi-task learning with information-theoretical metrics, properties of the label distributions: entropy (indicating the amount of uncertainty in the distribution) and kurtosis (indicating the skewness of the distribution). They have shown that the multi-task learning setup works best for tasks which have label distributions with lower kurtosis and relatively high entropy.   
Table \ref{tab6} shows the kurtosis and entropy properties of the label distribution for each of the events in the dataset, as well as token-type ratio (TTR).
Each of the events has very different properties; there is a strong difference in properties between larger events (top) and smaller ones (bottom). Smaller events tend to have more extreme values, which may explain that their addition to the evaluation adds further complexity to the task of rumour verification. 
In line with findings of Alonso and Plank \shortcite{alonso2017multitask}, in our case both auxiliary tasks have on average lower kurtosis than the main task. The stance classification dataset has on average higher entropy than the rumour detection dataset.

Whereas Alonso and Plank \shortcite{alonso2017multitask} were considering low-level linguistically related tasks as auxiliary tasks, such as part-of-speech tagging, we are working with higher level theme-related tasks. Therefore, it is interesting that we still observe similar trends when looking at the same data properties.
\begin{table}[t]
\centering
\resizebox{0.8\textwidth}{!}{%
\begin{tabular}{|l|l|l|l|l|l|l|l|l|l|}
\hline
\textbf{}                  & \multicolumn{3}{l|}{\textbf{Kurtosis}} & \multicolumn{3}{l|}{\textbf{Entropy}} & \multicolumn{3}{l|}{\textbf{TTR}}    \\ \hline
\textbf{Events}            & \textbf{S}  & \textbf{V}  & \textbf{D} & \textbf{S}  & \textbf{V} & \textbf{D} & \textbf{S} & \textbf{V} & \textbf{D} \\ \hline
\textbf{charliehebdo}      & -0.73       & -1.25       & -0.18      & 0.89        & 1.08       & 0.53       & 0.2        & 0.11       & 0.07       \\ \hline
\textbf{ottawashooting}    & -0.83       & 0.33        & -1.99      & 1.04        & 0.82       & 0.69       & 0.19       & 0.11       & 0.09       \\ \hline
\textbf{germanwings-crash} & -1.11       & -0.86       & -1.99      & 0.99        & 0.99       & 0.69       & 0.26       & 0.16       & 0.13       \\ \hline
\textbf{sydneysiege}       & -0.79       & 0.71        & -1.91      & 1.01        & 0.76       & 0.68       & 0.19       & 0.11       & 0.07       \\ \hline
\textbf{ferguson}          & -0.5        & 17.44       & -0.64      & 0.99        & 0.28       & 0.56       & 0.17       & 0.09       & 0.06       \\ \hline \hline
\textbf{ebola-essien}      & -0.22       & -3          & -3         & 1.01        & 0          & 0          & 0.38       & 0.27       & 0.27       \\ \hline
\textbf{putinmissing}      & -1.55       & 9.08        & -1.98      & 1.12        & 0.26       & 0.69       & 0.49       & 0.26       & 0.24       \\ \hline
\textbf{prince-toronto}    & -1.05       & 27.75       & 53.26      & 1.04        & 0.14       & 0.09       & 0.36       & 0.18       & 0.18       \\ \hline
\textbf{gurlitt}           & -           & 25.5        & -1.95      & -           & 0.14       & 0.68       & -          & 0.31       & 0.25       \\ \hline
\end{tabular}
}
\caption{Properties of the datasets for each of the events and each of the tasks, where S - Stance, V - Veracity, and D - Detection.}
\label{tab6}
\vspace{-0.5cm}
\end{table}
\section{Conclusions and Future Work}
We have proposed a rumour verification model that achieves improved performance for veracity classification by leveraging task relatedness with auxiliary tasks, specifically rumour detection and stance classification, through a multi-task learning approach. We have compared single task learning approaches with the proposed multi-task learning approaches that combine the verification classifier with the stance and rumour detection classifiers individually, as well as with both the rumour detection system and the stance classifier. Our results show that the joint learning of two tasks from the verification pipeline outperforms a single-learning approach to rumour verification. The combination of all three tasks leads to further performance improvements. We have also investigated the link between the properties of the label distribution in the dataset and the outcomes of our multi-task learning models. Our results support findings from previous research \cite{alonso2017multitask}. 

In future work we plan to investigate whether further improvements on main and auxiliary tasks are possible with multi-task learning by: adapting the training schedule to account for different dataset sizes, such that none of the tasks dominates the model; by incorporating the hierarchy between tasks into the model. We would like to investigate the effect of adding extra features (such as user features and interactions) on different tasks and at which stage to incorporate them, in private or shared layers.

\section*{Acknowledgements}
This work was supported by The Alan Turing Institute under the EPSRC grant EP/N510129/1. Cloud computing resources were kindly provided through a Microsoft Azure for Research Award. Work by Elena Kochkina was partially supported
by the Leverhulme Trust through the Bridges Programme and Warwick CDT for Urban Science \& Progress under the EPSRC Grant Number EP/L016400/1. 

\bibliographystyle{acl}
\bibliography{coling2018}

\end{document}